\documentclass[letterpaper, 10 pt, conference]{ieeeconf}
\IEEEoverridecommandlockouts
\usepackage{amsmath,amsfonts}
\usepackage{colortbl}
\usepackage[dvipsnames]{xcolor} 
\usepackage{algorithmic}
\usepackage{algorithm}
\usepackage{array}
\newcolumntype{C}[1]{>{\centering\arraybackslash}p{#1}}
\newcolumntype{L}[1]{>{\raggedright\arraybackslash}p{#1}}
\usepackage[caption=false,font=normalsize,labelfont=sf,textfont=sf]{subfig}
\usepackage{textcomp}
\usepackage{stfloats}
\usepackage{url}
\usepackage{verbatim}
\usepackage{graphicx}
\usepackage{pifont}
\usepackage{wrapfig}
\usepackage{cite}
\usepackage{amssymb}  
\usepackage{booktabs}  
\usepackage{arydshln}  
\usepackage{needspace}  
\usepackage{marvosym}  
\usepackage{hhline}
\usepackage{tabularx}
\usepackage{multirow} 
\usepackage{hyperref}

\begin{document}

\title{\LARGE \bf
CEDex: Cross-Embodiment Dexterous Grasp Generation at Scale from Human-like Contact Representations
}


\author{
Zhiyuan Wu$^1$, 
Rolandos Alexandros Potamias$^2$, 
Xuyang Zhang$^1$,
Zhongqun Zhang$^3$,
Jiankang Deng$^2$, 
Shan Luo$^1$
\thanks{$^1$Zhiyuan Wu, Xuyang Zhang, and Shan Luo are with Department of Engineering, King's College London, Strand, London, WC2R 2LS, United Kingdom, \{zhiyuan.1.wu, xuyang.zhang, shan.luo\}@kcl.ac.uk. }
\thanks{$^2$Rolandos Alexandros Potamias and Jiankang Deng are with Imperial College London, London, SW7 2AZ, United Kingdom, \ r.potamias@imperial.ac.uk, jiankangdeng@gmail.com. }
\thanks{$^3$Zhongqun Zhang is with College of Software, Nankai University, Tianjin, 300350, China, zhongqunzhang@outlook.com. }
}
\maketitle

\begin{abstract} 
Cross-embodiment dexterous grasp synthesis refers to adaptively generating and optimizing grasps for various robotic hands with different morphologies. This capability is crucial for achieving versatile robotic manipulation in diverse environments and requires substantial amounts of reliable and diverse grasp data for effective model training and robust generalization. However, existing approaches either rely on physics-based optimization that lacks human-like kinematic understanding or require extensive manual data collection processes that are limited to anthropomorphic structures. In this paper, we propose CEDex, a novel cross-embodiment dexterous grasp synthesis method at scale that bridges human grasping kinematics and robot kinematics by aligning robot kinematic models with generated human-like contact representations. Given an object's point cloud and an arbitrary robotic hand model, CEDex first generates human-like contact representations using a Conditional Variational Auto-encoder pretrained on human contact data. It then performs kinematic human contact alignment through topological merging to consolidate multiple human hand parts into unified robot components, followed by a signed distance field-based grasp optimization with physics-aware constraints. Using CEDex, we construct the largest cross-embodiment grasp dataset to date, comprising 500K objects across four gripper types with 20M total grasps. Extensive experiments show that CEDex outperforms state-of-the-art approaches and our dataset benefits cross-embodiment grasp learning with high-quality diverse grasps.
\href{https://georgewuzy.github.io/cedex-website/}{Project Page}.
\end{abstract}

\section{Introduction} 
Humans, by nature, possess remarkable dexterous grasping capabilities that can generate feasible and reliable grasps for given objects while seamlessly adapting to various constraints and generalizing across different finger configurations, \textit{e.g.}, three or four-fingered grasps \cite{liu2023contactgen}. In robotic manipulation systems, such a capability of performing versatile grasping is equally important, as grasping is fundamental for complex downstream tasks \cite{she2024crosshandpolicies}. However, most existing robotic grasping methods are tailored to specific end-effectors, which hampers their generalization. When faced with new robotic hands featuring novel morphologies, these approaches necessitate costly data collection and time-consuming retraining for each embodiment, which limits their practical deployment and scalability \cite{li2023gendexgrasp}. This limitation indicates the urgent need for a unified model capable of representing and generating grasps across various robotic embodiments. Such a capability, which enables the adaptive generation and optimization of grasps for arbitrary robotic hands, is referred to as \textbf{cross-embodiment} grasp synthesis \cite{wei2024drograsp}. 
While recent efforts \cite{attarian2023geomatch, wei2024geomatch++, wei2024drograsp} have demonstrated the potential of using unified representations to learn cross-embodiment dexterous grasping, training these models requires substantial amounts of reliable and diverse data. The data need to include grasps that satisfy both physical constraints and human-like kinematic plausibility, quantified through metrics such as force closure stability and dynamic simulation success rates. Additionally, these learning-based methods often face data imbalance, with certain grasp orientations being overrepresented while others remain underexplored. Therefore, it is crucial to develop a robust cross-embodiment grasp data synthesis strategy at scale. 

\begin{figure}
    \centering
    \vspace{-0.6em}
    \includegraphics[width=0.99\linewidth]{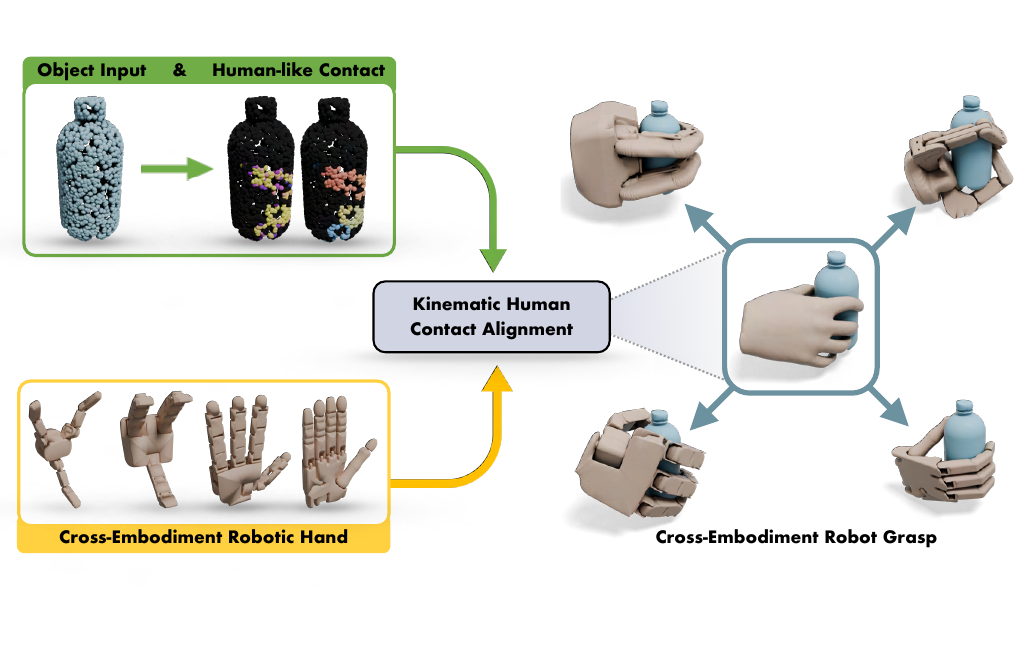}
    \vspace{-4.2em} 
    \caption{Given an object point cloud and the generated human-like contact representations, our proposed CEDex method can generate dexterous grasps across various robotic hand embodiments. By integrating both human-like kinematics and physics-aware constraints, CEDex is able to synthesize stable and diverse grasp at scale.}
    \label{fig:teaser}
    \vspace{-1.0em} 
\end{figure}

Existing cross-embodiment grasp synthesis methods for large-scale data generation primarily rely on grasp optimization, utilizing physical constraints such as force closure to ensure grasp feasibility \cite{liu2021dfc, li2023gendexgrasp}. However, these physics-based approaches focus solely on static equilibrium and neglect human grasping kinematics, failing to consider the dynamic nature of grasping process, where the hand must plausibly approach and engage with the object. This oversight leads to low success rates in practical scenarios. To incorporate dynamic kinematic considerations, some approaches have attempted to learn from human demonstrations \cite{chao2021dexycb, liu2024realdex}, but this process requires significant manual effort for data collection, leading to increased costs. Additionally, these methods often necessitate calibrating gripper joints and remapping them to corresponding human hand joints, limiting their effectiveness primarily to anthropomorphic structures \cite{she2024crosshandpolicies}. As a result, non-anthropomorphic robotic hands, such as the three-fingered Barrett and Robotiq-3F, still face significant challenges.

Recent advancements in human grasp synthesis have shown that model-based grasp optimization can effectively simulate how a human hand approaches and grasps an object \cite{liu2023contactgen, grady2021contactopt, jiang2021grasptta}. These methods leverage contact representations to model grasping at the object-centric level and utilize learning-based models for generation, a process that is significantly more efficient than collecting and retargeting from human grasp demonstrations. Inspired by this, we propose CEDex, a novel \underline{\textbf{C}}ross-\underline{\textbf{E}}mbodiment \underline{\textbf{Dex}}terous grasp synthesis method at scale that aligns robot kinematic models with generated human-like contact representations during the grasping process. Specifically, given a point cloud of an object and the kinematic model of a robotic hand, CEDex consists of two key components: human contact generation and kinematic human contact alignment. Once a human-like contact is generated from a Conditional Variational Auto-encoder (CVAE) model pretrained on human contact data, the kinematic human contact alignment component conducts topological merging to integrate multiple human hand parts into cohesive robot components, aligned with the target robot's kinematic configuration. Subsequently, a signed distance function (SDF)-based grasp optimization with physics-aware constraints is employed to produce robust and diverse grasps that reflect human-like kinematic understanding. Using CEDex, we constructed a large-scale cross-embodiment dexterous grasp dataset, which contains 500K objects and four types of grippers, with 20M grasps in total. Extensive experiments have demonstrated the effectiveness and superiority of our proposed method and dataset over state-of-the-art (SoTA) approaches.

Our main contributions can be summarized as follows: 
\begin{itemize}
    \item We propose CEDex, a novel cross-embodiment dexterous grasp synthesis method at scale that aligns robot kinematic models and generated human-like contact representations to enable effective grasp optimization across diverse robotic hands.
    \item We construct the largest cross-embodiment grasp dataset to the best of our knowledge, with 500K objects, four types of grippers, and 20M grasps in total.
    \item Extensive experiments demonstrate that our proposed CEDex outperforms SoTA cross-embodiment grasp synthesis approaches. Furthermore, our constructed large-scale dataset provides high-quality diverse grasps that significantly benefit cross-embodiment grasp learning.
\end{itemize}

\section{Related Works}

\subsection{Robotic Dexterous Grasping}
Dexterous grasping acts as an essential element for various complex, human-like manipulation tasks. In recent years, data-driven approaches have emerged as a promising direction for dexterous grasping. Shao \textit{et al.} has pioneered UniGrasp \cite{shao2020unigrasp} for multi-fingered robotic hand grasping by learning generalizable contact point representations. Subsequent works such as GraspTTA \cite{jiang2021grasptta} and ContactGen \cite{liu2023contactgen} employ contact maps as intermediate representations to effectively bridge hand-object geometry and synthesize diverse grasps through optimization-based approaches. The UniDexGrasp series \cite{xu2023unidexgrasp, wan2023unidexgrasp++} further introduce universal policies capable of handling thousands of object instances via sophisticated curriculum learning and teacher-student distillation frameworks. Building upon this trend, recent research has shifted from single-embodiment, object-agnostic approaches to both hand-agnostic and object-agnostic methodologies, marking a significant advancement toward truly universal grasping systems. The GeoMatch series \cite{attarian2023geomatch, wei2024geomatch++} and the recent DRO-Grasp \cite{wei2024drograsp} achieve cross-embodiment dexterous grasp synthesis by learning geometric correspondences between diverse hand morphologies and object geometries. This development highlights the critical importance of robust and diverse grasp data, as well as the need for grasp synthesis methods at scale. 

\subsection{Dexterous Grasp Datasets}
Generating dexterous grasp data is foundational for data-driven dexterous grasping. Direct capture of human grasp demonstrations, as exemplified by works including GRAB \cite{taheri2020grab} and RealDex \cite{liu2024realdex}, provides the most behaviorally faithful and accurate supervision by retargeting human grasps to generate reliable grasp data. However, this reliance on manual capture and retargeting constrains dataset scale and object diversity. To reduce collection cost and expand coverage, other datasets synthesize grasps in simulator like GraspIt! \cite{miller2004graspit} and Isaac Gym \cite{liang2018isaacgym}. For instance, DDG \cite{liu2020ddg} synthesizes dexterous grasp poses in GraspIt! \cite{miller2004graspit}, and DexGraspNet \cite{wang2023dexgraspnet} generates grasps via differentiable force-closure with Isaac Gym physics validation. Recently, works such as GraspXL \cite{zhang2024graspxl} and DexGrasp Anything \cite{zhong2025dexgraspanything} further push scalable generation by synthesizing objective-conditioned grasp motions and large-scale dexterous grasps over extensive object sets, \textit{e.g.}, Objaverse \cite{deitke2023objaverse}. Moving beyond single anthropomorphic settings, the community is pivoting toward cross-embodiment and multi-object generalization of dexterous grasping. DFC \cite{liu2021dfc} first proposed to synthesizes grasps for arbitrary hand morphologies with a differentiable force-closure estimator. Based on it, MultiDex \cite{li2023gendexgrasp} offers a multi-hand dataset to support cross-hand generalization. Recently, Multi-GraspLLM \cite{li2024multigraspllm} leverages LLM-driven annotations to directly generate final grasp poses. However, current approaches still lack human-like kinematic understanding of grasp dynamics and remain limited at scale (both object diversity and grasp counts), underscoring the need for more robust, diverse, large-scale datasets with rich contact and kinematically informed representations across embodiments. 

\begin{figure*}
    \centering
    \vspace{0.6em}
    \includegraphics[width=0.8\linewidth]{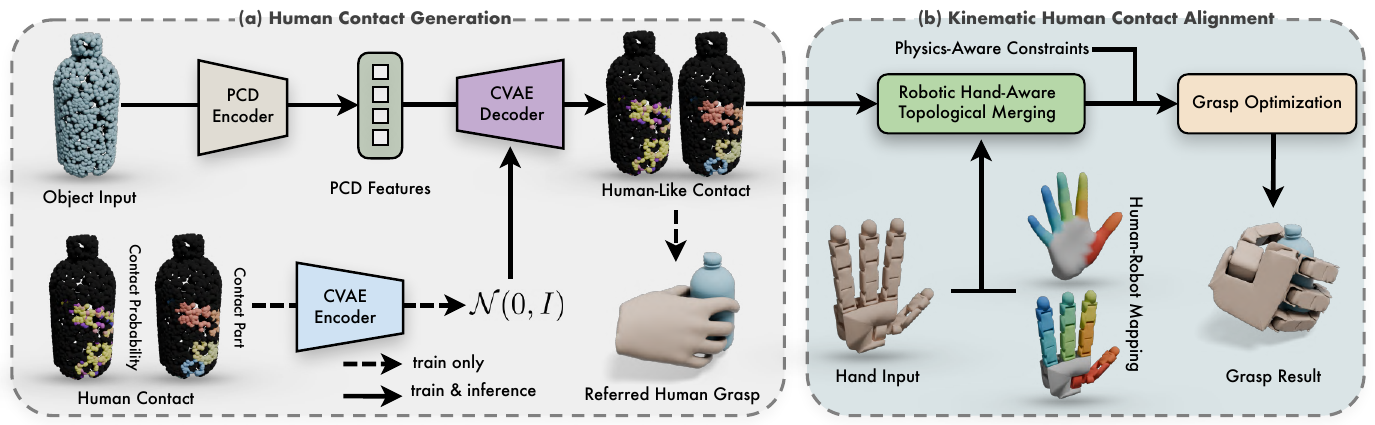}
    \vspace{-1em}
    \caption{
    The cross-embodiment dexterous grasp synthesis pipeline of our CEDex. Given a point cloud of an object and a robotic hand as input, CEDex first generates human-like contact representations using a CVAE model pretrained on human contact data. The kinematic human contact alignment component then performs topological merging to consolidate multiple human hand parts into unified robot components according to the robot's kinematic configuration, followed by a SDF-based grasp optimization with physics-aware constraints to generate robust and diverse grasps with human-like kinematic understanding.
    }
    \vspace{-1.0em} 
    \label{fig:pipeline}
\end{figure*}

\section{Generating Cross-Embodiment Grasp from Human Contact}
As illustrated in Fig. \ref{fig:pipeline}, CEDex takes 1) the point cloud $\boldsymbol{O} \in \mathbb{R}^{N\times3}$ of an object, where $N$ represents point number, and 2) the kinematic model of an arbitrary robotic hand as input, and generates a physically stable grasp configuration for the given robotic hand. The architecture consists of two key components: a) a human contact generation part that generates reliable human-like contact representations using a CVAE model pretrained on human contact data, and b) a kinematic human contact alignment part that performs topological merging to consolidate multiple human hand parts into unified robot components according to the target robot's kinematic configuration, followed by signed distance field-based grasp optimization with physics-aware constraints to generate robust and diverse grasps with human-like kinematics. The details of each component are elaborated below.

\subsection{Human Contact Generation}
We first generate human-like contact representations using a pretrained CVAE model. We employ a MANO hand \cite{romero2017mano} to represent human hand and divide it into $B=16$ parts, referring to \cite{liu2023contactgen}. The object-centric human contact representations $[\boldsymbol{C}^h, \boldsymbol{P}^h]$ consist of a contact map $\boldsymbol{C}^h \in \mathbb{R}^{N \times 1}$ and a part map $\boldsymbol{P}^h \in \mathbb{R}^{N \times B}$. Each contact value $\boldsymbol{c}_k \in [0, 1]$ in $\boldsymbol{C}^h$ represents the contact probability of point $\boldsymbol{o}_k$ in $\boldsymbol{O}$. The definition and computation of $\boldsymbol{C}^h$ follows ContactOpt \cite{grady2021contactopt}, where a virtual capsule is placed at each object point $\boldsymbol{o}_k$, and $\boldsymbol{c}_k$ is set to 1 if any point in the human hand point cloud lies inside the capsule and otherwise smoothly decays with distance. The one-hot part map $\boldsymbol{P}^h$ represents which part of the hand touches the object, indicating the hand part label in $\{1, \dots, B\}$ in contact with each object point. The direction map from \cite{liu2023contactgen} is not considered, as the substantial shape differences between human fingers and robotic gripper fingers result in significantly different contact directions, making direct human-to-robot transfer ineffective. We employ CVAE to model the conditional probabilities $p(\boldsymbol{C}^h, \boldsymbol{P}^h|\boldsymbol{O})$ as: 
\begin{equation}
    p(\boldsymbol{C}^h, \boldsymbol{P}^h|\boldsymbol{O}) = p(\boldsymbol{P}^h|\boldsymbol{C}^h, \boldsymbol{O})p(\boldsymbol{C}^h|\boldsymbol{O}),
\end{equation}
where $\boldsymbol{C}^h$ is conditioned on object $\boldsymbol{O}$ and $\boldsymbol{P}^h$ is additionally conditioned on $\boldsymbol{C}^h$. We use two Point Cloud \cite{qi2017pointnet} decoders $\mathcal{D}_c$ and $\mathcal{D}_p$ to predict $\boldsymbol{C}^h$ and $\boldsymbol{P}^h$ as:
\begin{equation}
    \boldsymbol{C}^h = \mathcal{D}_c (\boldsymbol{z}_c, \boldsymbol{F}^o),
\end{equation}
and
\begin{equation}
    \boldsymbol{P}^h = \mathcal{D}_p (\boldsymbol{z}_p, \boldsymbol{C}^h, \boldsymbol{F}^o), 
\end{equation}
where $\boldsymbol{F}^o$ represents feature maps extracted from $\boldsymbol{O}$ via PointNet++ \cite{qi2017pointnet++}, and $\boldsymbol{z}_c \sim \mathcal{N} (\boldsymbol{0}, \boldsymbol{I})$ and $\boldsymbol{z}_p \sim \mathcal{N} (\boldsymbol{0}, \boldsymbol{I})$ represent latent codes sampled from Gaussian distributions. We pretrain our human contact generation model with:
\begin{equation}
\mathcal{L}^{recon} = |\boldsymbol{C}^h - \hat{\boldsymbol{C}}^h| + \lambda_p\mathcal{L}_{CE}(\boldsymbol{P}^h, \hat{\boldsymbol{P}}^h) + \lambda_{KL}\mathcal{L}^{KL},
\end{equation}
where $|\boldsymbol{C}^h - \hat{\boldsymbol{C}}^h|$ is the L1 loss for contact maps, $\mathcal{L}_{CE}(\boldsymbol{P}^h, \hat{\boldsymbol{P}}^h)$ is the cross-entropy loss for part maps, and a KL regularization loss $\mathcal{L}^{KL}$ \cite{kingma2013vae}. The model is pretrained on the GRAB \cite{taheri2020grab} and YCB Affordance \cite{corona2020ycbaffordance} datasets. 

\subsection{Kinematic Human Contact Alignment} 
Directly transferring human-like contact representations to robotic hands poses significant challenges due to fundamental structural differences between human hands and robotic grippers. These differences manifest in two key aspects: 1) varying finger numbers across different robotic embodiments, and 2) distinct joint configurations within a single finger. Such morphological disparities make direct keypoint remapping or grasp optimization with raw human-like contact representations unavailable for most robotic hands. However, humans demonstrate remarkable adaptability in performing stable grasps using different finger combinations, where multiple fingers can collectively fulfill the mechanical role of a single finger across different configurations. This observation inspires our approach to perform topological merging of contact representations at the object level. We adaptively consolidate multiple human hand parts, such as the ring and pinky fingers, into fewer robotic components, such as a single gripper finger, based on the target robot's kinematic configuration. This strategy facilitates effective kinematic alignment between human grasping mechanics and the capabilities of robotic hands, enabling robots to perform human-like kinematics in grasping.

\begin{figure}
    \centering
    \includegraphics[width=0.99\linewidth]{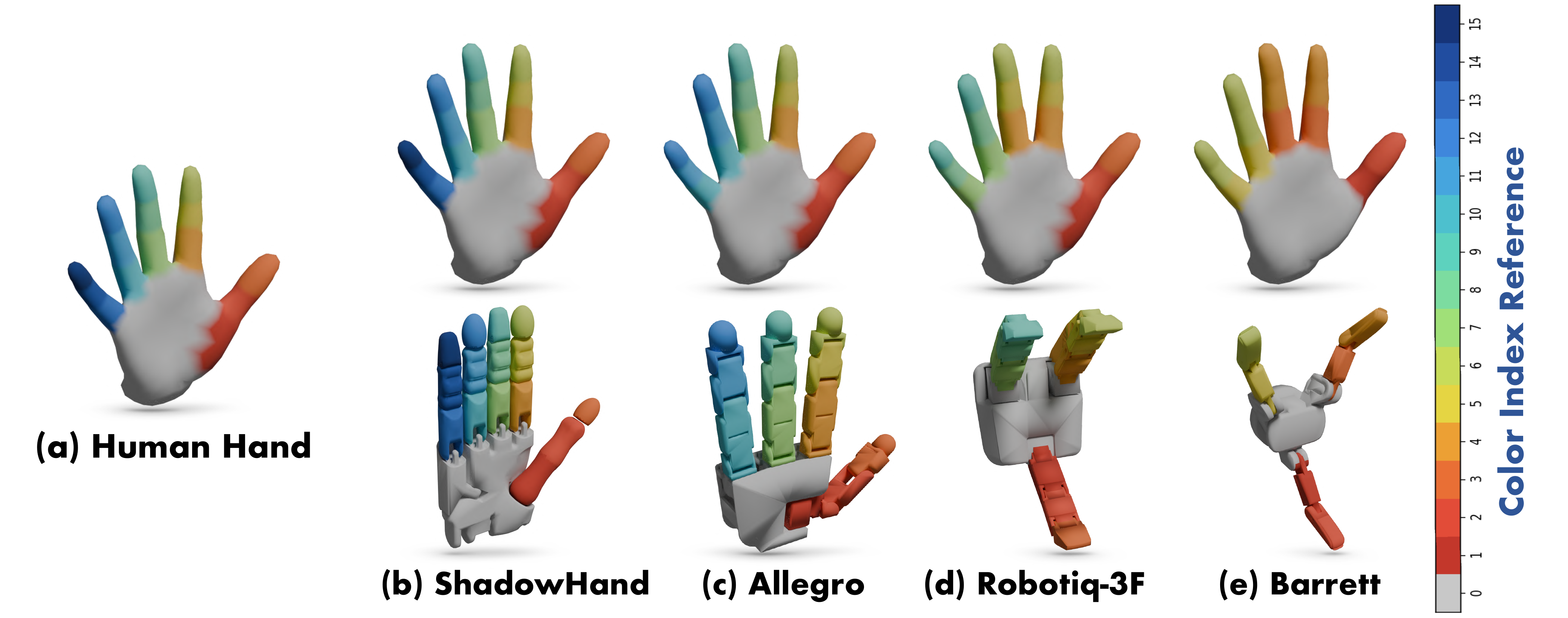}
    \vspace{-1.8em} 
    \caption{Human-robot mapping to align human hand parts with robotic hand components. (a) Original human hand (b) Shadow hand (c) Allegro (d) Robotiq-3F (e) Barrett. Color index reference for visualization is provided. }
    \vspace{-1.0em} 
    \label{fig:hand_remap}
\end{figure}

We predefine a kinematic human-robot mapping to align human hand parts with robotic hand components according to their configurations, as shown in Fig. \ref{fig:hand_remap}. Since human hands have more complex part structures than robotic hands, this mapping requires merging multiple human parts into single robot parts. 
To implement this predefined mapping, we develop a geometric-based topological merging approach that consolidates contact representations from multiple human parts into unified robot contact representations at the object level. Given human-like contact representations $[\boldsymbol{C}^h, \boldsymbol{P}^h]$ with contact map $\boldsymbol{C}^h \in \mathbb{R}^{N \times 1}$ and part map $\boldsymbol{P}^h \in \mathbb{R}^{N \times B}$, our goal is to generate robot-specific contact representations $[\boldsymbol{C}^r, \boldsymbol{P}^r]$, where $\boldsymbol{C}^r \in \mathbb{R}^{N \times 1}$ is the robot contact map and $\boldsymbol{P}^r \in \mathbb{R}^{N \times B'}$ is the robot part map with $B'$ robot components. When human parts $\{b_i, b_j\} \subset \{1, ..., B\}$ need to be merged into a single robot component $b_m \in \{1, ..., B'\}$, we first extract the two contact parts $\boldsymbol{P}^i = \boldsymbol{P}^h[:,\ i]$ and $\boldsymbol{P}^j = \boldsymbol{P}^h[:,\ j]$. We then obtain the corresponding part-specific contact maps through element-wise multiplication as:
\begin{equation}
\boldsymbol{C}^i = \boldsymbol{C}^h \odot \boldsymbol{P}^i, \quad \boldsymbol{C}^j = \boldsymbol{C}^h \odot \boldsymbol{P}^j,
\end{equation}
where $\boldsymbol{C}^i, \boldsymbol{C}^j \in \mathbb{R}^{N \times 1}$ represent the contact values for human parts $i$ and $j$ respectively, and $\odot$ denotes element-wise multiplication. 
For each point $\boldsymbol{o}_x$ with its contact value $\boldsymbol{c}^i_x > 0$, the projection direction is designed to consolidate the spatially separated contacts from two human parts into a unified region that can be effectively accessed by a single robot component, where the direction towards the centroid between the two parts ensures optimal geometric coverage for the merged contact. We compute the projecting direction as:
\begin{equation}
\boldsymbol{v}_x = \frac{\frac{\overrightarrow{\boldsymbol{M}^o \boldsymbol{o}_x}}{|\overrightarrow{\boldsymbol{M}^o \boldsymbol{o}_x}|} + \frac{\overrightarrow{\boldsymbol{M}^o \boldsymbol{M}^j}}{|\overrightarrow{\boldsymbol{M}^o \boldsymbol{M}^j}|}}{\left\|\frac{\overrightarrow{\boldsymbol{M}^o \boldsymbol{o}_x}}{|\overrightarrow{\boldsymbol{M}^o \boldsymbol{o}_x}|} + \frac{\overrightarrow{\boldsymbol{M}^o \boldsymbol{M}^j}}{|\overrightarrow{\boldsymbol{M}^o \boldsymbol{M}^j}|}\right\|},
\end{equation}
where $\boldsymbol{M}^o$ represents the object mass centroid and $\boldsymbol{M}^i$ and $\boldsymbol{M}^j$ represent the mass centroids of $\boldsymbol{C}^i$ and $\boldsymbol{C}^j$, computed as:
\begin{equation}
\boldsymbol{M}^o = \frac{1}{N}\sum_{k=1}^{N}\boldsymbol{o}_k, 
\end{equation}
\begin{equation}
    \boldsymbol{M}^{i,j} = \frac{\sum_{k=1}^{N} \boldsymbol{c}^{i,j}_k \cdot \boldsymbol{o}_k}{\sum_{k=1}^{N} \boldsymbol{c}^{i,j}_k},
\end{equation}
where $\boldsymbol{c}^i_k$ and $\boldsymbol{c}^j_k$ represent the contact value of $k$-th point in $\boldsymbol{C}^i$ and $\boldsymbol{C}^j$, and $\boldsymbol{o}_k$ represent the $k$-th point in the object point cloud $\boldsymbol{O}$. The remapped position ${\boldsymbol{o}_x}'$ is then determined as the nearest point in $\boldsymbol{O}$ to the ray from $\boldsymbol{M}^o$ in direction $\boldsymbol{v}_x$. By applying the same remapping process to all points in human part $b_j$ towards $b_i$, we obtain the symmetric remapped contact pair. Next, given the two remapped contact pairs $[{\boldsymbol{C}^i}', {\boldsymbol{P}^i}']$ and $[{\boldsymbol{C}^j}', {\boldsymbol{P}^j}']$ with ${\boldsymbol{C}^i}'$, ${\boldsymbol{C}^j}'$, ${\boldsymbol{P}^i}'$, ${\boldsymbol{P}^j}' \in \mathbb{R}^{N\times1}$, the merged contact $[\boldsymbol{C}^m, \boldsymbol{P}^m]$ is obtained by taking the union of the two remapped contact pairs as:
\begin{equation}
\boldsymbol{c}^m_k = \begin{cases}
{\boldsymbol{c}^i_k}' + {\boldsymbol{c}^j_k}' & \text{if } {\boldsymbol{p}^i_k}' > 0 \text{ and } {\boldsymbol{p}^j_k}' > 0 \\
{\boldsymbol{c}^i_k}' & \text{if } {\boldsymbol{p}^i_k}' > 0 \text{ and } {\boldsymbol{p}^j_k}' = 0 \\
{\boldsymbol{c}^j_k}' & \text{if } {\boldsymbol{p}^i_k}' = 0 \text{ and } {\boldsymbol{p}^j_k}' > 0 \\
0 & \text{otherwise},
\end{cases}
\end{equation}
where ${\boldsymbol{p}^i_k}'$ and ${\boldsymbol{p}^j_k}'$ represent the remapped part assignment values at point $k$ for parts $i$ and $j$ respectively, and ${\boldsymbol{c}^i_k}'$ and ${\boldsymbol{c}^j_k}'$ are the corresponding remapped contact values. For cases where more than two human parts need to be merged into a single robot component, we repeat this pairwise merging process iteratively. 

Following the predefined kinematic human-robot mapping, we apply the above merging operations to all required human part combinations to generate the robot-specific contact representations $[\boldsymbol{C}^r, \boldsymbol{P}^r]$. We then employ signed distance field-based contact optimization for human-robot contact alignment. The contact loss $\mathcal{L}_c$ is formulated as:
\begin{equation}
\mathcal{L}_c = \sum_{k=1}^{N} \boldsymbol{c}^r_k \sum_{b=1}^{B'} \boldsymbol{p}^r_{k,b} \cdot |\text{SDF}_b(\boldsymbol{o}_k)|, 
\end{equation}
where $\boldsymbol{c}^r_k$ is the contact value at object point $\boldsymbol{o}_k$, $\boldsymbol{p}^r_{k,b}$ indicates the assignment of point $\boldsymbol{o}_k$ to robot part $b$, and $\text{SDF}_b(\boldsymbol{o}_k)$ represents the signed distance from robotic hand part $b$ to object point $\boldsymbol{o}_k$.

\begin{table*}[!t]
    \centering\footnotesize{
    \vspace{0.6em}
	\caption{
            Comparison of dexterous grasp datasets. Our dataset achieves the largest scale on cross-embodiment dexterous hands to the best of our knowledge, with considerations of both human-like kinematics and kinematic physical awareness. 
    }
\label{tab:dataset_comparison}
    \vspace{-0.8em} 
    \begin{tabular}{l c c c c c}
    \toprule
        \multirow{2}{*}{Dataset} & \multirow{2}{*}{Hand Type} & \multirow{2}{*}{Object} & \multirow{2}{*}{Grasp} & Human-Like & Kinematic Physical \\
        & & & & Kinematics & Awareness \\
        \hline
        GRAB \cite{taheri2020grab} & Human & 51 & 1.64M & \textcolor{ForestGreen}{\ding{51}} & \textcolor{BrickRed}{\ding{55}} \\
        DexYCB \cite{chao2021dexycb} & Human & 20 & 1K & \textcolor{ForestGreen}{\ding{51}} & \textcolor{BrickRed}{\ding{55}} \\
        DDG \cite{liu2020ddg} & Single & 565 & 6.9K & \textcolor{BrickRed}{\ding{55}} & \textcolor{ForestGreen}{\ding{51}} \\
        DexGraspNet \cite{wang2023dexgraspnet} & Single & 5.3K & 1.32M & \textcolor{BrickRed}{\ding{55}} & \textcolor{ForestGreen}{\ding{51}} \\
        UniDexGrasp \cite{xu2023unidexgrasp} & Single & 5.5K & 1.12M & \textcolor{BrickRed}{\ding{55}} & \textcolor{ForestGreen}{\ding{51}} \\
        DexGrasp Anything \cite{zhong2025dexgraspanything} & Single & 15.6K & 3.4M & \textcolor{BrickRed}{\ding{55}} & \textcolor{ForestGreen}{\ding{51}} \\
        RealDex \cite{liu2024realdex} & Single & 59K & 52 & \textcolor{ForestGreen}{\ding{51}} & \textcolor{BrickRed}{\ding{55}} \\
        GraspXL \cite{zhang2024graspxl} & Anthropomorphic & \textbf{500K} & - & \textcolor{BrickRed}{\ding{55}} & \textcolor{BrickRed}{\ding{55}} \\
        \hdashline
        MultiDex \cite{li2023gendexgrasp} & \textbf{Cross-Embodiment} & 58 & 436K & \textcolor{BrickRed}{\ding{55}} & \textcolor{BrickRed}{\ding{55}} \\
        Multi-GraspLLM \cite{li2024multigraspllm} & \textbf{Cross-Embodiment} & 2.1K & 140K & \textcolor{BrickRed}{\ding{55}} & \textcolor{BrickRed}{\ding{55}} \\
        \textbf{CEDex (Ours)} & \textbf{Cross-Embodiment} & \textbf{500K} & \textbf{20M} & \textcolor{ForestGreen}{\ding{51}} & \textcolor{ForestGreen}{\ding{51}} \\
    \bottomrule
    \end{tabular}
    \vspace{-1.0em} 
}\end{table*}

\begin{figure}
    \centering
    \vspace{0.6em}
    \includegraphics[width=0.65\linewidth]{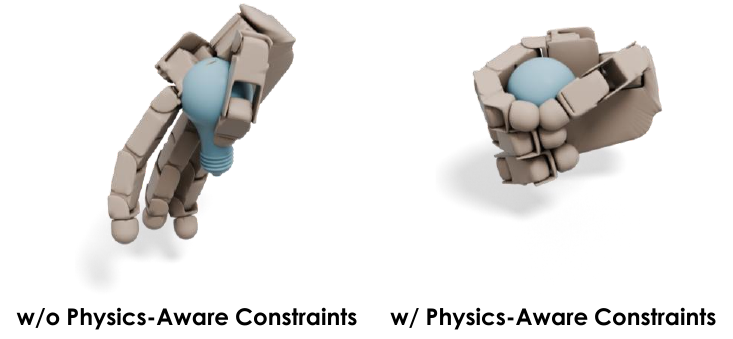}
    \vspace{-1em} 
    \caption{An example of Allegro hand grasping a light bulb, demonstrating the robustness and stability provided by physics-aware constraints, including surface pulling forces and prevention of penetration. }
    \vspace{-1.0em} 
    \label{fig:physics_aware_constraints}
\end{figure}

\subsection{Physics-Aware Constraints} 

While human grasping primarily relies on natural kinematic principles, robot grasp synthesis requires additional stability guarantees. Therefore, beyond the contact loss $\mathcal{L}_c$ in human-robot contact alignment, we follow \cite{zhong2025dexgraspanything} and incorporate additional physics-aware constraints to enhance the robustness of synthesized grasps. We incorporate three key physical constraints: Surface Pulling Force (SPF) loss \cite{xu2023unidexgrasp} that encourages proximity between robot parts and the object surface, External-penetration Repulsion Force (ERF) loss \cite{li2023gendexgrasp} that prevents hand-object collisions, and Self-penetration Repulsion Force (SRF) loss \cite{xu2023unidexgrasp} that maintains realistic hand geometry by preventing finger self-intersections:
\begin{equation}
\mathcal{L}_{SPF} = \frac{\sum_{k \in \mathcal{S}} \sqrt{d_k^o}}{|\mathcal{S}| + \eta},
\end{equation}
\begin{equation}
\mathcal{L}_{ERF} = \frac{1}{B'} \sum_{b=1}^{B'} \max_{k \in \mathcal{H}_b} |\min(0, \text{SDF}_{obj}(\boldsymbol{h}_k))|,
\end{equation}
and
\begin{equation}
\mathcal{L}_{SRF} = \frac{1}{B'} \sum_{b=1}^{B'} \sum_{i,j \in \mathcal{H}_b, i \neq j} \max(0, d_{th} - ||\boldsymbol{h}_i - \boldsymbol{h}_j||),
\end{equation}
where $d_k^o$ represents the distance from robotic hand point $\boldsymbol{h}_k$ to the object surface, $\mathcal{S}$ is the set of hand points within threshold distance, $\mathcal{H}_b$ denotes the set of hand points belonging to robot part $b$, $\text{SDF}_{obj}(\boldsymbol{h}_k)$ represents the signed distance from the object surface to hand point $\boldsymbol{h}_k$, and $d_{th}$ is the self-collision threshold distance. An example is showing in Fig. \ref{fig:physics_aware_constraints}. 

\section{CEDex dataset}
Using the proposed CEDex method for synthesizing robust and diverse cross-embodiment dexterous grasps, we construct a large-scale cross-embodiment dexterous grasp dataset. For robotic hand selection, we choose four diverse robotic hands ranging from three to five fingers: Barrett Hand, Robotiq-3F, Allegro, and Shadow hand. We exclude two-finger grippers as we do not regard them as dexterous robotic hands, and two-finger grasps cannot achieve stable multi-directional force application and fail to pass rigorous multi-directional stability tests \cite{li2023gendexgrasp}. For object selection, we utilize 58 real-world objects from ContactDB \cite{brahmbhatt2019contactdb} and YCB \cite{calli2015ycb} datasets following \cite{li2023gendexgrasp}, and 503,409 synthesized objects from the large-scale Objaverse \cite{deitke2023objaverse} dataset following \cite{zhang2024graspxl}.

We generate our dataset using eight 32GB NVIDIA Tesla V100 GPUs. We first pretrain the human contact generation CVAE model using human grasp data from GRAB \cite{taheri2020grab} and YCB Affordance \cite{corona2020ycbaffordance} datasets on the 58 real-world objects. Then for each real-world object, we generate 64 human-like contact pairs consisting of contact maps and part maps. For each contact pair, we perform grasp optimization with 64 randomly sampled initial gripper wrist poses, which ensures grasp diversity even for each single human-like contact pairs, and select the top-16 grasps with the lowest optimization energy scores. This process yields 59,392 grasp candidates for each robotic hand, totaling 237,568 grasps, which is comparable to the scale of MultiDex \cite{li2023gendexgrasp}. To significantly expand the dataset scale, we process the 500K synthesized Objaverse objects with our CEDex. For each Objaverse object, we generate 4 human-like contact pairs, and for each contact pair, we perform grasp optimization with 16 randomly sampled initial gripper wrist poses, selecting the top-4 grasps with the lowest energy scores. This results in 8M grasp candidates for each robotic hand, totaling 32M grasps across all hands. We then apply rigorous filtering using Isaac Gym \cite{liang2018isaacgym} based on comprehensive stability evaluation metrics detailed in Sec. \ref{sec:exp_setup}. After filtering, our final dataset contains 20M high-quality grasps. We provide both generated grasping poses and final grasping poses after Isaac Gym validation in our dataset. 

As demonstrated in Tab. \ref{tab:dataset_comparison}, our CEDex dataset is the largest cross-embodiment dexterous grasp dataset to date in terms of both object diversity and grasp quantity. Crucially, we are the first dataset to simultaneously incorporate 1) human-like kinematics that leverages natural human grasping principles and contact patterns during the whole grasp synthesis process and 2) kinematic physical awareness that enforces physical constraints of hand kinematics, addressing limitations of existing datasets that typically focus on only one aspect.

\section{Experiments}

\subsection{Experimental Setup} \label{sec:exp_setup}
We evaluate our CEDex framework on an 80GB NVIDIA A100 GPU. Following the evaluation protocol established in \cite{li2023gendexgrasp}, we conduct experiments on a test set comprising 10 representative unseen daily objects from the ContactDB \cite{brahmbhatt2019contactdb} and YCB \cite{calli2015ycb} datasets. Our evaluation spans three diverse robotic hands with varying finger configurations: Barrett (3 fingers), Allegro (4 fingers), and Shadow hand (5 fingers). Note that we do not evaluate baseline methods on Robotiq-3F as they do not support this gripper, though we provide CEDex results on it in our analysis. 

For implementation, we pretrain our CVAE model to generate human-like contact representations using training parameters from \cite{liu2023contactgen} on 58 real-world objects from the MultiDex dataset \cite{li2023gendexgrasp}. Subsequently, we perform grasp optimization over 200 iterations for final grasp synthesis.

We provide comprehensive comparisons against established cross-embodiment dexterous grasp synthesis baselines, including optimization-based methods DFC \cite{liu2021dfc} and GenDexGrasp \cite{li2023gendexgrasp}, as well as learning-based approaches GeoMatch \cite{wei2024geomatch++}, GeoMatch++ \cite{wei2024geomatch++}, and DRO-Grasp \cite{wei2024drograsp}. To evaluate grasp synthesis performance, we employ success rate and diversity, defined as:
\begin{itemize}
    \item \textbf{Success Rate}: We evaluate grasping success by applying external forces to the object and measuring its displacement. Using the Isaac Gym simulator \cite{liang2018isaacgym}, a simple grasp controller executes the predicted grasps \cite{wei2024drograsp}. Following the metric definition in \cite{li2023gendexgrasp}, we sequentially apply forces along six orthogonal directions for 1 second each. A grasp is considered successful if the object's displacement remains below 2 cm once all forces are applied. 

    \item \textbf{Diversity}: Grasp diversity is quantified by computing the standard deviation of joint configurations across all successful grasps, including the 6-DoF wrist pose and finger joint angles. Higher standard deviation indicates greater diversity in the generated grasp configurations. 

\end{itemize}

\begin{table*}[t!]  
    \centering  
    \vspace{0.6em}
    \caption{Quantitative results of our CEDex compared with different cross-embodiment dexterous grasp synthesis baselines across three robotic hands from three to five fingers: Barrett, Allegro, and Shadow hand. We evaluate success rate and diversity. }  
    \label{tab:grasp_comparison}  
    \vspace{-0.8em} 
    \begin{tabular}{l | c c c c | c c c c}
        \toprule  
        \multirow{2}{*}{Method} & \multicolumn{4}{c}{Success Rate (\%) $\uparrow$} & \multicolumn{4}{c}{Diversity (rad.) $\uparrow$} \\
        \cmidrule(lr){2-5} \cmidrule(lr){6-9}
        & Barrett & Allegro & ShadowHand & \textbf{Average} & Barrett & Allegro & ShadowHand & \textbf{Average} \\
        \midrule
        DFC \cite{liu2021dfc} & 86.3 & 76.2 & 58.8 & 73.8 & \cellcolor{LimeGreen!50}0.532 & \cellcolor{LimeGreen!50}0.454 & 0.435 & \cellcolor{LimeGreen!50}0.474 \\
        GenDexGrasp \cite{li2023gendexgrasp} & 67.0 & 51.0 & 54.2 & 57.4 & 0.488 & 0.389 & 0.318 & 0.398 \\
        GeoMatch \cite{attarian2023geomatch} & 60.0 & - & 67.5 & 63.8 &  0.259 & - &  0.235 & 0.247 \\
        GeoMatch++ \cite{wei2024geomatch++} & 77.5 & - & 70.0 & 73.8 &  0.378 & - &  0.184 & 0.281 \\
        DRO-Grasp \cite{wei2024drograsp} & \cellcolor{LimeGreen!50}87.3 & \cellcolor{LimeGreen}\textbf{92.3} & \cellcolor{LimeGreen!50}83.0 & \cellcolor{LimeGreen!50}87.5 & 0.513 & 0.397 & \cellcolor{LimeGreen}\textbf{0.441} & 0.450 \\
        \hline
        \textbf{CEDex (Ours)} & \cellcolor{LimeGreen}\textbf{93.1} & \cellcolor{LimeGreen!50}88.1 & \cellcolor{LimeGreen}\textbf{85.0} & \cellcolor{LimeGreen}\textbf{88.7} & \cellcolor{LimeGreen}\textbf{0.624} & \cellcolor{LimeGreen}\textbf{0.473} & \cellcolor{LimeGreen!50}0.438 & \cellcolor{LimeGreen}\textbf{0.512} \\
        \bottomrule  
    \end{tabular}  
    \vspace{-1.0em} 
\end{table*}

\subsection{Comparison with SoTAs}
As shown in Tab. \ref{tab:grasp_comparison}, our quantitative results reveal distinct trade-offs between different approaches. Learning-based cross-embodiment methods (GeoMatch \cite{attarian2023geomatch}, GeoMatch++ \cite{wei2024geomatch++}, DRO-Grasp \cite{wei2024drograsp}) generally achieve higher success rates than optimization-based approaches (DFC \cite{liu2021dfc}, GenDexGrasp \cite{li2023gendexgrasp}), particularly on complex-structural robotic hands like Shadow hand. However, these data-driven methods suffer from limited diversity due to training data imbalances, where certain grasp orientations may be overrepresented in the dataset while others remain underexplored. This limitation is evident from diversity metrics, where optimization-based methods demonstrate superior diversity compared to learning-based counterparts. 

\begin{figure*}
    \centering
    \includegraphics[width=0.9\linewidth]{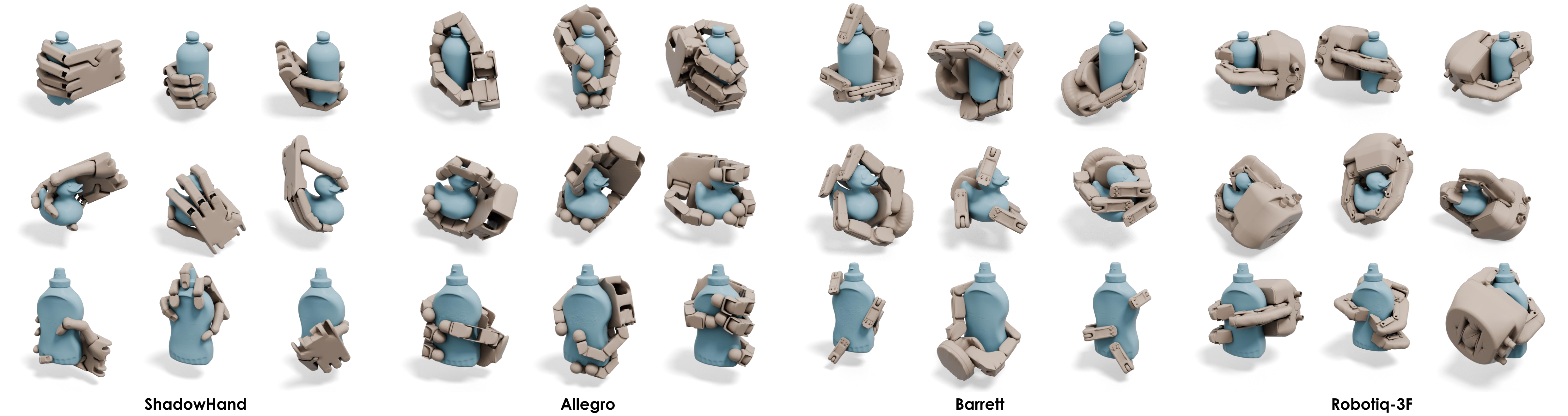}
    \vspace{-1em}
    \caption{
    Visualization of grasp results synthesized by our proposed CEDex, where our method generates robust and diverse grasps. 
    }
    \vspace{-1.0em} 
    \label{fig:vis_results}
\end{figure*}

In contrast, CEDex achieves the best performance across both success rate (88.7\%) and diversity (0.512 rad.). This superior performance comes from two key advantages. First, CEDex improves grasp success rates with human-like kinematics. Unlike optimization-based methods that rely solely on physical constraints, our approach leverages kinematic human contact alignment by employing a learning-based CVAE model to generate kinematically grounded human-like contact representations with functional understanding that enables better performance on complex hand structures. Compared to optimization-based methods (DFC \cite{liu2021dfc}, GenDexGrasp \cite{li2023gendexgrasp}), our CEDex improves success rates by 26.2\% to 30.8\% on Shadow hand. In addition, CEDex improves grasp diversity through data-agnostic optimization. It avoids training on robotic grasp datasets and performs grasp optimization from spatially uniform initial poses around the object, which eliminates the aforementioned data imbalance problems. Compared to learning-based methods (GeoMatch \cite{attarian2023geomatch}, GeoMatch++ \cite{wei2024geomatch++}, DRO-Grasp \cite{wei2024drograsp}), our CEDex improves average diversity by 12.1\% to 51.8\%. It is noteworthy that while Robotiq-3F results are not included in Tab. I due to the unavailability of baseline comparisons, CEDex achieves strong performance on it with \textbf{91.9\%} success rate of and \textbf{0.401} diversity. 

To evaluate the practical applicability of CEDex, we measure the computational time and GPU memory requirements for complete grasp synthesis, from 3D object and gripper input to final grasp pose generation. CEDex achieves remarkable efficiency with an average inference time of \textbf{7.8s} and \textbf{684 MiB} GPU memory per synthesis session, generating 64 grasps at \textbf{0.12s} per grasp. Compared to existing methods (batch size 64), CEDex demonstrates substantial speedup: DFC \cite{liu2021dfc} requires 1800s per batch (230 times slower), GenDexGrasp \cite{li2023gendexgrasp} needs 19.7s per batch (2.5 times slower), and DRO-Grasp \cite{wei2024drograsp} takes 0.65s per single grasp with 4 GB GPU memory (batch size 1) (5.4 times slower per grasp), highlighting CEDex's superior computational efficiency for practical robotic applications.

\subsection{Grasp Synthesis Visualization} 

\begin{figure}
    \centering
    \includegraphics[width=0.9\linewidth]{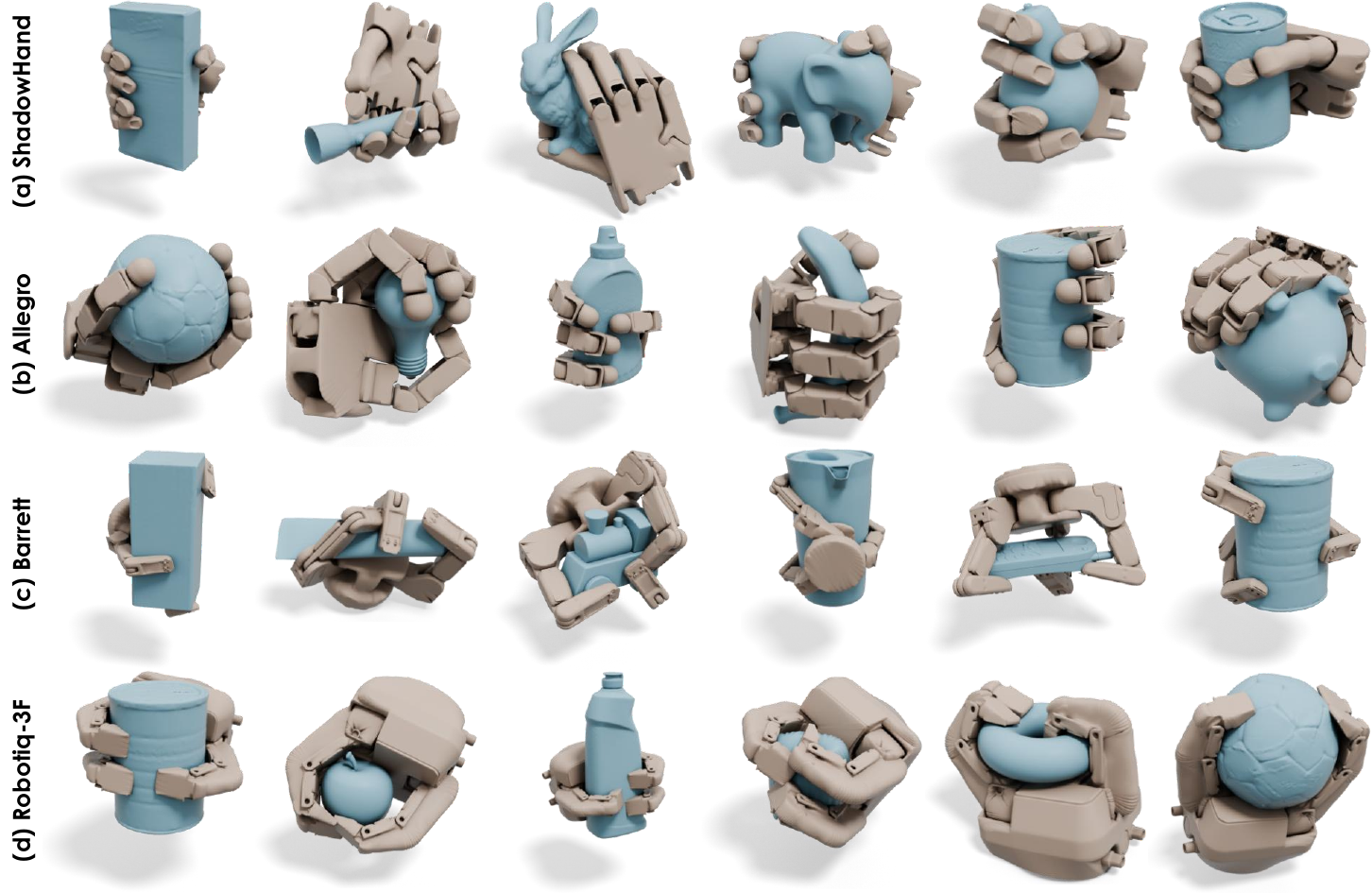}
    \vspace{-1em} 
    \caption{Visualization of more grasp results synthesized by our CEDex. }
    \vspace{-1.0em} 
    \label{fig:more_results}
\end{figure}

We provide qualitative results of our CEDex in Fig. \ref{fig:vis_results}, where our method generates robust and diverse grasps. It is worth mentioning that the multiple grasps shown for each object-hand pair in the figure are derived from a single generated human-like contact representation using different initial wrist poses. This illustrates how our method achieves grasp diversity by varying the wrist configuration while balancing human kinematic understanding and physical feasibility through physics-aware constraints. As a result, different initial wrist poses lead to diverse grasp outcomes, even from the same generated human-like contact representation. More results are available in Fig. \ref{fig:more_results} and our project repository. 

\begin{table*}[t!]  
    \centering  
    \vspace{0.6em}
    \caption{Quantitative results of DRO-Grasp \cite{wei2024drograsp} trained on our CEDex dataset compared with its dataset baseline across three robotic hands from three to five fingers: Barrett, Allegro, and Shadow hand. We evaluate success rate and diversity. }
    \label{tab:train_dro}  
    \vspace{-0.8em} 
    \begin{tabular}{l | c c c c | c c c c}
        \toprule  
        \multirow{2}{*}{Dataset} & \multicolumn{4}{c}{Success Rate (\%) $\uparrow$} & \multicolumn{4}{c}{Diversity (rad.) $\uparrow$} \\
        \cmidrule(lr){2-5} \cmidrule(lr){6-9}
        & Barrett & Allegro & ShadowHand & \textbf{Average} & Barrett & Allegro & ShadowHand & \textbf{Average} \\
        \midrule
        MultiDex \cite{li2023gendexgrasp} & 87.3 & 92.0 & 83.0 & 87.5 & 0.513 & 0.397 & \textbf{0.441} & 0.450 \\
        \textbf{CEDex (Ours)} & \textbf{91.2} & \textbf{95.4} & \textbf{86.4} & \textbf{91.0} & \textbf{0.524} & \textbf{0.419} & 0.436 & \textbf{0.460} \\
        \bottomrule  
    \end{tabular}  
    \vspace{-1.0em} 
\end{table*}

\subsection{Training Learning-based Networks on CEDex dataset}
To validate the effectiveness of our proposed CEDex dataset, we train the learning-based cross-embodiment grasp synthesis network DRO-Grasp \cite{wei2024drograsp} using our CEDex dataset. For direct comparison of data quality, we evaluate on the ContactDB and YCB object sets, which have the same scale as the MultiDex dataset \cite{li2023gendexgrasp} originally used by DRO-Grasp. The results are reported in Tab. \ref{tab:train_dro}, where DRO-Grasp trained on our CEDex dataset achieves superior metrics across all metrics except for diversity on Shadow hand when compared to the baseline trained on MultiDex. It shows 3.5\% and 2.2\% improvement on success rate and diversity, demonstrating the effectiveness and superiority of our constructed dataset for cross-embodiment grasp learning. 

\subsection{Ablation Study} 

\begin{table}[t!]  
    \centering  
    \caption{
    Ablation study on kinematic human contact alignment on four robotic hands from three to five fingers: Barrett, Robotiq-3F, Allegro, and Shadow hand. We report average success rate and diversity. 
    }  
    \label{tab:ablation_contact}  
    \vspace{-0.8em} 
    \begin{tabular}{c c c}
        \toprule  
        Method & Success Rate (\%) $\uparrow$ & Diversity (rad.) $\uparrow$ \\
        \midrule
        w/o Alignment & 27.7 & 0.372 \\
        w/ Alignment & \textbf{89.3} & \textbf{0.484} \\
        \bottomrule  
    \end{tabular}  
\end{table}

\textbf{Kinematic Human Contact Alignment}. We evaluate our kinematic human contact alignment against the hand-agnostic contact loss from \cite{li2023gendexgrasp} as our baseline, where both approaches incorporate our physics-aware constraints to ensure fair comparison. Quantitative results are reported in Tab. \ref{tab:ablation_contact}, where our method with kinematic alignment achieves a success rate of 89.3\% (a 61.1\% improvement) and a diversity score of 0.484 (a 23.1\% improvement). This substantial improvement stems from our kinematic alignment's ability to effectively adapt human-like contact representations to robot-specific morphologies, while the hand-agnostic baseline neglects critical embodiment differences that are essential for successful cross-embodiment transfer. 

\begin{table}[t!]  
    \centering  
    \caption{
    Ablation study on kinematic physics-aware constraints on four robotic hands from three to five fingers: Barrett, Robotiq-3F, Allegro, and Shadow hand. We report average success rate and diversity.
    }  
    \label{tab:ablation_physics}  
    \vspace{-0.8em} 
    \begin{tabular}{c c c c c}
        \toprule  
        SPF & ERF & SRF & Success Rate (\%) $\uparrow$ & Diversity (rad.) $\uparrow$ \\
        \midrule
        - & - & - & 30.9 & 0.413 \\
        \ding{51} & - & - & 86.7 & 0.405 \\
        \ding{51} & \ding{51} & - & 87.2 & 0.412 \\
        \ding{51} & \ding{51} & \ding{51} & \textbf{89.3} & \textbf{0.484} \\
        \bottomrule  
    \end{tabular}  
    \vspace{-1.0em} 
\end{table}

\textbf{Physics-Aware Constraints}. We systematically evaluate the contribution of our physics-aware constraints through incremental ablation studies. Starting from the proposed method without any constraints, we progressively add the three physics-aware constraints. As reported in Tab. \ref{tab:ablation_physics}, each constraint contributes to the overall performance, especially the surface pulling loss (SPF) \cite{xu2023unidexgrasp}, which enhances the success rate from 30.9\% to 86.7\% with a 55.8\% improvement. The results underscores the importance of physics-aware constraints in bridging the gap between kinematic alignment and physical realizability, ensuring that human-inspired grasps remain feasible when executed on diverse robotic embodiments. 

\subsection{Real-World Validation}

\begin{wrapfigure}[8]{r}{0.23\textwidth}
    \centering
    \vspace{-2.2em}
    \includegraphics[width=0.22\textwidth]{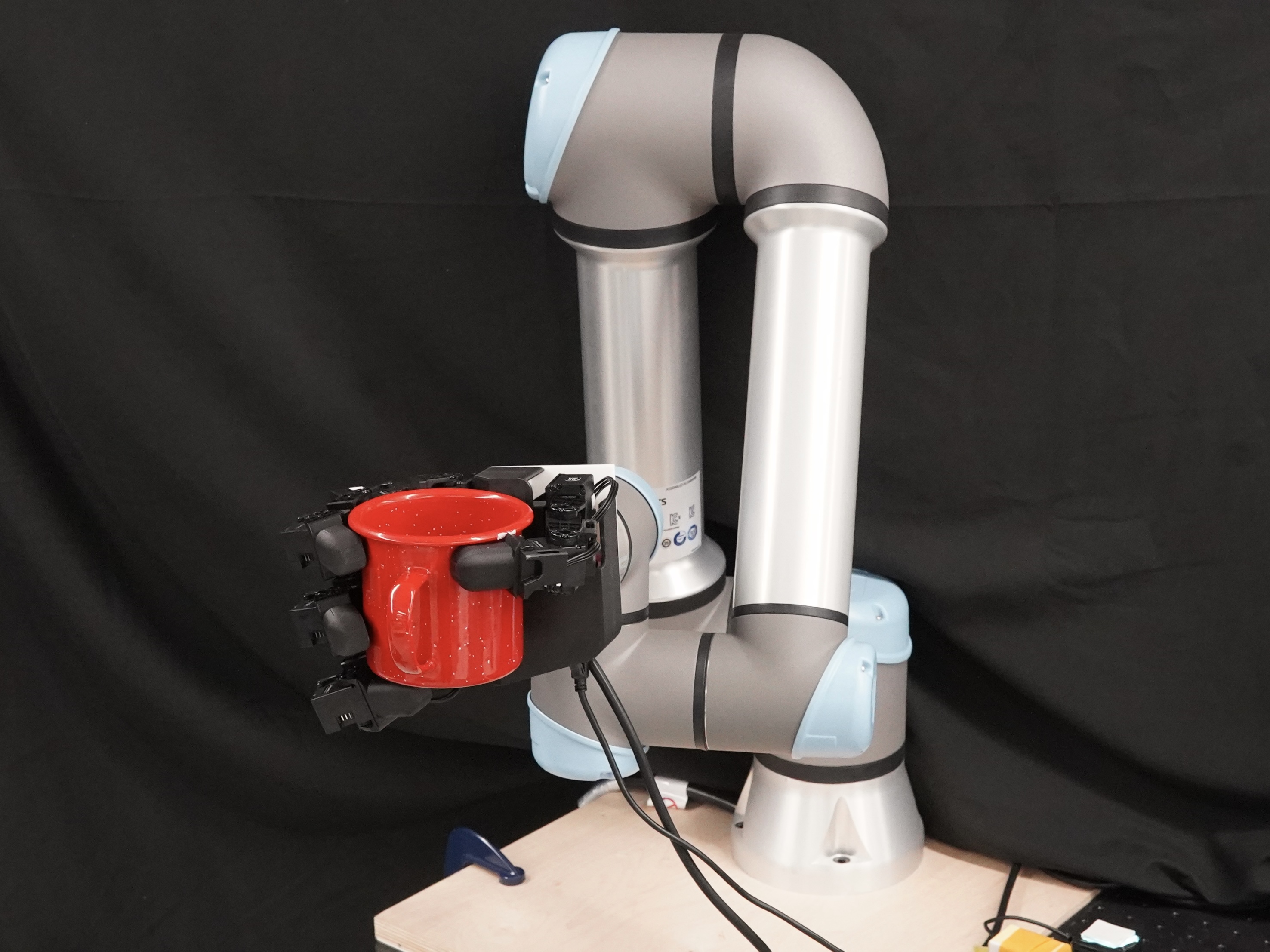}
    \vspace{-0.6em}
    \caption{Our real world setting. }
    \label{fig:real}
\end{wrapfigure}

We validate our CEDex in real world using a UR5-e robot equipped with a Leap Hand \cite{shaw2023leaphand}, as shown in Fig. \ref{fig:real}, showcasing its effectiveness in dexterous grasping and generalization to previously unseen objects. More results and videos are available in our project repository.

\section{Conclusion and Discussion}
In this paper, we present CEDex, a novel cross-embodiment dexterous grasp synthesis method that bridges human grasping kinematics and robot kinematics through kinematic human contact alignment. By generating human-like contact representations and performing topological merging followed by SDF-based optimization, CEDex enables large-scale synthesis of physically feasible and kinematically plausible grasps across diverse robotic embodiments. We construct the largest cross-embodiment grasp dataset to date with 500K objects and 20M grasps across four gripper types, demonstrating superior performance over existing approaches. Extensive experiments validate the effectiveness of our approach and demonstrate that our method and dataset significantly advance the state-of-the-art in cross-embodiment dexterous grasping. 

In this work, we exclude two-finger grippers from our framework as they are not considered as dexterous robotic hands, and two-finger grasps are inherently unable to achieve stable multi-directional force application and consistently fail to pass the rigorous multi-directional stability tests \cite{li2023gendexgrasp}, which are essential for our comprehensive stability evaluation framework. For future work, we will focus on leveraging our large-scale dataset to develop advanced learning-based approaches, including diffusion models and other foundation model architectures, to further enhance cross-embodiment dexterous grasping capabilities.

{
\tiny  
\bibliographystyle{IEEEtran}
\bibliography{main}
}

\end{document}